\documentclass[letterpaper]{article} 
\usepackage{aaai2027} 
\usepackage[hyphens]{url}  
\usepackage{graphicx} 
\usepackage{natbib}  
\usepackage{caption} 
\usepackage{algorithm}
\usepackage{algpseudocode}
\usepackage{graphicx}
\usepackage{subcaption}
\usepackage{newfloat}
\usepackage{listings}
\usepackage{amsmath}
\usepackage{amssymb}
\usepackage{amsfonts}
\DeclareCaptionStyle{ruled}{labelfont=normalfont,labelsep=colon,strut=off} 
\floatstyle{ruled}
\newfloat{listing}{tb}{lst}{}
\floatname{listing}{Listing}

\usepackage{booktabs}

\title{One Teacher, Many Views: Reducing Reference Sensitivity in On-Policy Self-Distillation}

\author{
    Khouloud Saadi\corresponding,
    Di Wang
}
\affiliations{
    \textsuperscript{} Department of Computer Science\\

    King Abdullah University of Science and Technology\\
    KAUST, Thuwal, Saudi Arabia\\
}

\begin{document}

\maketitle
\begin{abstract}
On-policy self-distillation (OPSD) improves reasoning by training a model on its own trajectories with dense token-level supervision from a privileged copy conditioned on a reference solution. Existing methods typically use a single teacher distribution conditioned on the complete reference. We identify a limitation of this formulation: a reference is only one realization of the reasoning process. Although it provides evidence about the answer, it also encodes an ordering, phrasing, and solution path. Full-reference conditioning can therefore produce correct targets that depend excessively on information unavailable to the student at inference. We show that replacing one valid reference with another, while holding the problem and student state fixed, can substantially alter the teacher distribution, a phenomenon we call reference-induced target instability. To address this instability, we introduce Monte Carlo Privileged Distillation (MCPD), which models privileged context as a distribution over partial reference views and distills the marginal teacher distribution. MCPD evaluates the teacher under multiple views and averages the resulting next-token distributions. We show that this probability-space average is the exact target induced by minimizing the expected teacher-to-student KL divergence over privileged views. Across mathematical and scientific reasoning benchmarks, we show that MCPD produces more reference-consistent supervision and consistently improves over full-reference OPSD.

\end{abstract}

\section{Introduction}
\label{sec:introduction}
\begin{figure*}[t]
    \centering

    \begin{subfigure}[t]{0.38\textwidth}
        \centering
        \includegraphics[width=\linewidth]
        {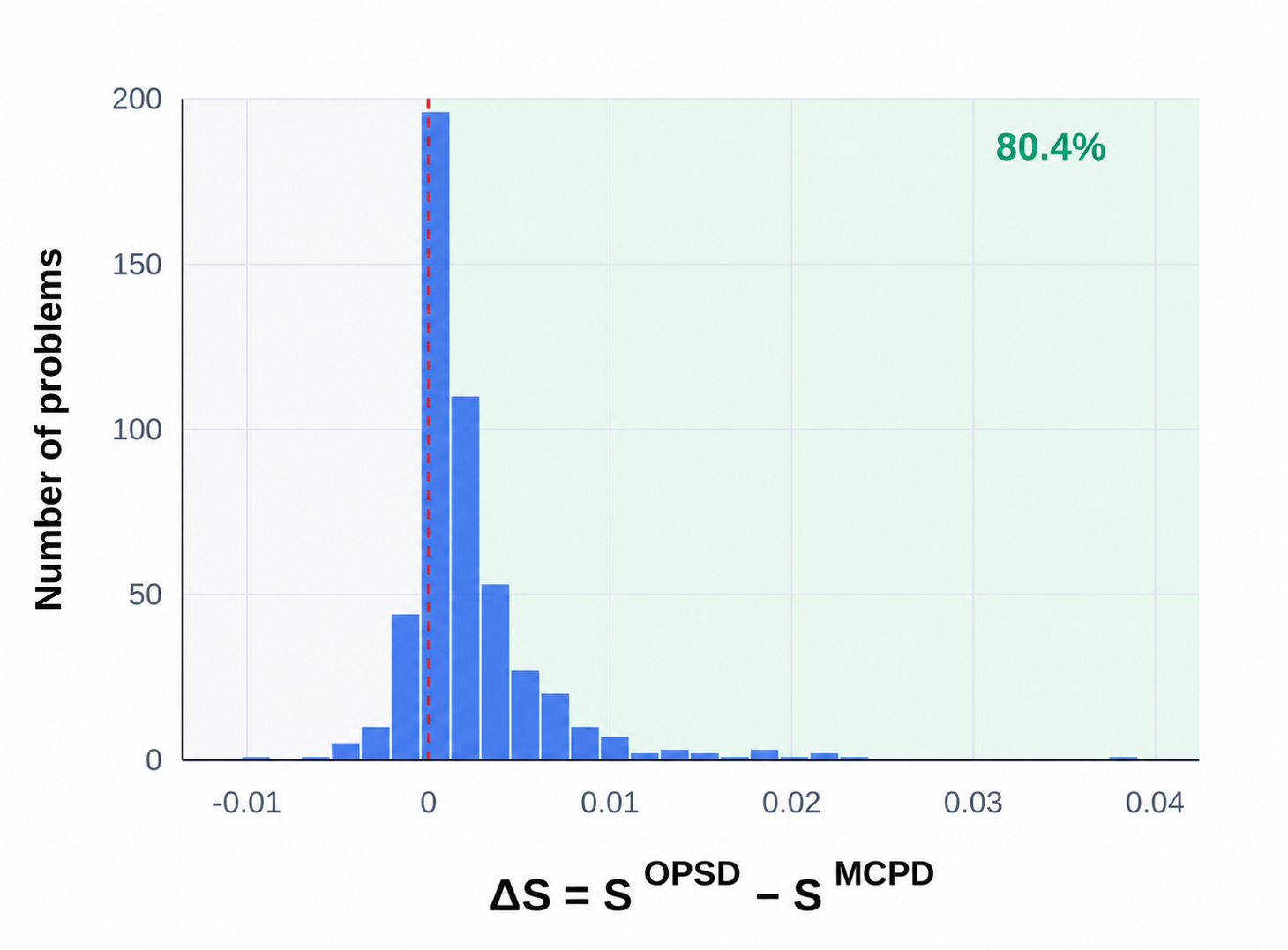}
    \end{subfigure}
    \hfill
    \begin{subfigure}[t]{0.6\textwidth}
        \centering
        \includegraphics[width=\linewidth]{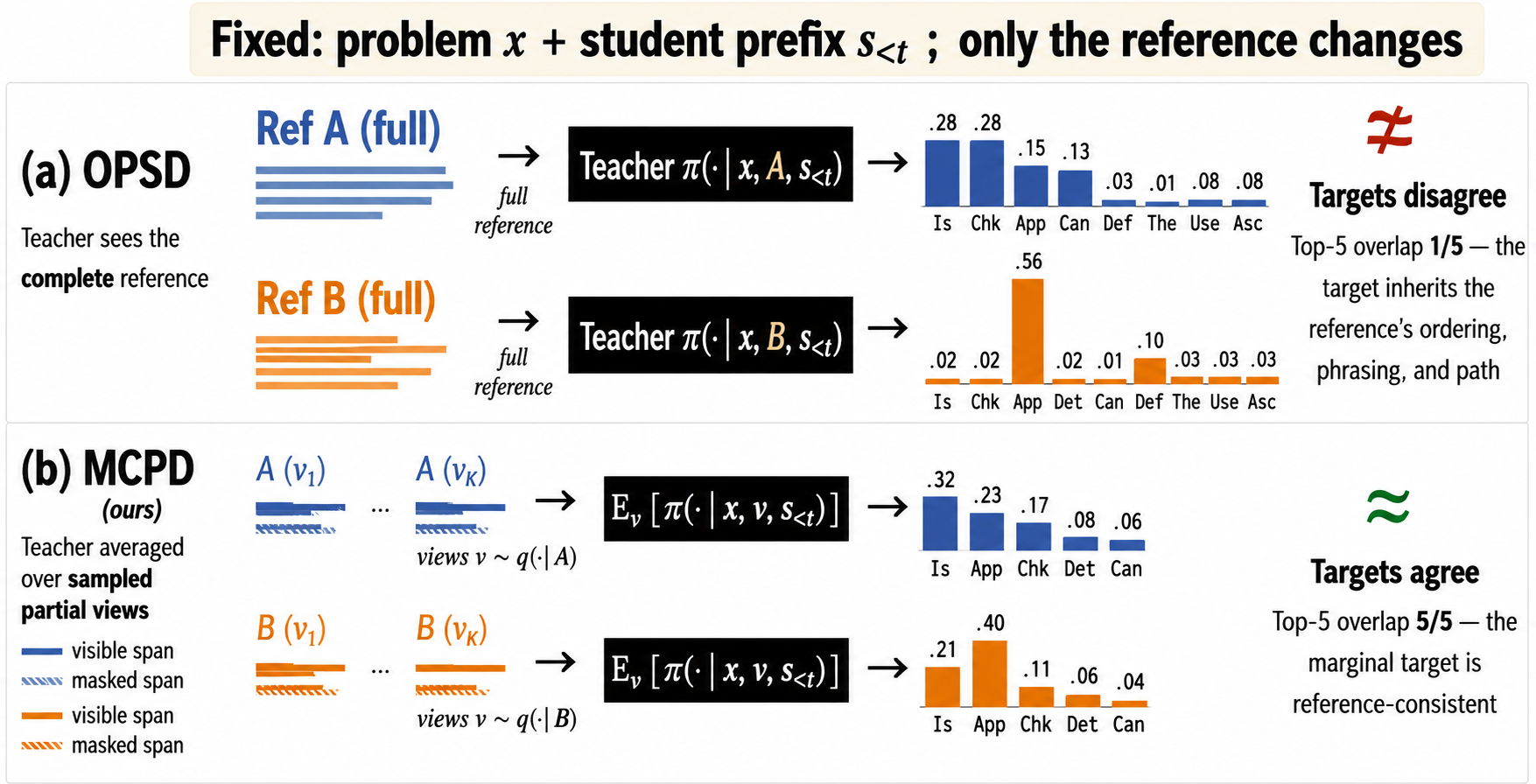}
    \end{subfigure}
    \caption{
Reference-induced target instability. \textbf{Left:} Distribution of the per-problem sensitivity
reduction
\(\Delta S=S^{\mathrm{OPSD}}-S^{\mathrm{MCPD}}\) (defined in Method)
over \(500\) problems sampled from
\texttt{DeepMath-103K}~\citep{he2026deepmath} using
\texttt{Qwen3-8B}. For each problem, we use two valid
reference solutions, generate one unprivileged student rollout,
and evaluate both methods at matched prefix positions along
the same trajectory. MCPD averages \(K=4\) partial views of
each reference; positive values indicate lower reference
sensitivity than OPSD.
\textbf{Right:} Qualitative example with the problem and
student prefix fixed, showing that full-reference OPSD
produces more reference-dependent next-token targets, whereas
MCPD yields more consistent marginal targets.
}
    \label{fig:reference_sensitivity}
\end{figure*}
Training strong reasoning models requires more than knowing
whether the final answer is correct; the model also needs
feedback on the intermediate reasoning states it actually
visits during generation~\citep{agarwal2024policy,song2026survey}. Supervised fine-tuning~\citep{yumetamath,zelikman2022star}
provides dense token-level guidance from reference solutions,
but trains the model on fixed expert trajectories rather than
on its own reasoning paths~\citep{guha2025openthoughts}.
Reinforcement learning~\citep{guo2025deepseek,yu2026dapo}
instead trains on model-generated trajectories, but typically
relies on rewards assigned to complete solutions, providing
limited guidance about individual reasoning steps. On-policy
distillation~\citep{saadi2026validity,agarwal2024policy,gu2024minillm} aims to combine these advantages by supplying
dense teacher feedback along trajectories generated by the
student itself.

On-Policy Self-Distillation (OPSD)~\citep{zhao2026self,
nguyen2026avsd,song2026survey} realizes this idea without
requiring a separate, stronger model. The student reasons from
the problem and its own generated prefix, whereas a
stop-gradient teacher additionally observes a verified
reference solution. At each student state, the teacher provides
a full next-token distribution that is distilled into the
student. This construction combines on-policy state coverage,
dense supervision, and privileged reasoning information
within a single model. However, we uncover a surprising instability in OPSD: the teacher target can change substantially even when the problem and student-generated prefix are held fixed and both privileged references are fully correct. This occurs because the teacher distribution depends not only on the problem and student state, but also on the particular reference solution supplied as privileged context~\citep{zhao2026self}. A reference is only one valid realization of the reasoning process, and alternative correct solutions may differ in their decomposition, intermediate quantities, notation, ordering, or reasoning path. Crucially, we find that slightly changing one valid reference can induce significantly different next-token distributions for exactly the same student state. Consequently, two equivalent references may prescribe conflicting learning updates, revealing that OPSD supervision can be highly sensitive to incidental properties of the chosen solution. We term this phenomenon \emph{reference-induced target instability}.

Figure~\ref{fig:reference_sensitivity} illustrates both an
example of this phenomenon and its prevalence. In
Figure~\ref{fig:reference_sensitivity}   (Right), the problem and
student prefix are fixed and only the valid reference supplied
to the teacher slightly changes. Full-reference OPSD produces sharply
different next-token targets, with only one shared token among
the two top-\(5\) sets. In contrast, our approach (MCPD) constructs a marginal
target from sampled partial views of each reference, producing
complete top-\(5\) overlap in the illustrated state.
Figure~\ref{fig:reference_sensitivity} (Left) shows that this is not
an isolated example: MCPD reduces reference sensitivity on $80.4\%$ of the
500 evaluated problems. For this comparison, we use the same pair of valid references
for OPSD and MCPD and hold the student trajectory and
evaluated states fixed. OPSD compares the targets induced by
the two complete references, whereas MCPD separately
constructs a marginal target from partial views of each
reference. 

This matched protocol ensures that differences between OPSD and
MCPD are not confounded by the problem, student trajectory,
evaluated positions, or choice of reference pair.

Importantly, MCPD's lower reference sensitivity is not explained by
target flattening. On $270$ problems from the same matched evaluation
set, MCPD produces lower full-vocabulary entropy than OPSD on
$75.6\%$ of problems, while their mean entropy remains comparable
($0.33$ vs.\ $0.34$). Analysis details are provided in Appendix.

Reference-induced instability arises because a reference
serves two roles simultaneously. It provides task-relevant
evidence about how the problem can be solved, but it also
exposes the teacher to one particular presentation of that
solution. Conditioning on the complete reference does not
distinguish between these two sources of information. The
resulting target may therefore encode both useful guidance and
reference-specific properties of a particular derivation. Our
goal is not to remove privileged information or force all
valid reasoning paths to induce identical predictions. Rather,
we seek a target that preserves the guidance supplied by the
reference while reducing dependence on any single complete
presentation of that guidance. This motivates the following:

\begin{center}
\fbox{%
\parbox{0.92\columnwidth}{%
\centering
\emph{Privileged supervision should reflect predictions that
remain stable across partial views of the reference.}
}%
}
\end{center}

We introduce \textbf{Monte Carlo Privileged
Distillation} (\textbf{MCPD}) based on this principle. Instead
of treating the complete reference as a single deterministic
privileged context, we define a distribution over its partial
views. Each view reveals a different subset of the same
solution and induces a corresponding teacher distribution. We
distill the expectation of these distributions rather than the
prediction produced by one complete view. For a fixed student
state \(s_t\), reference \(r\), and random partial-view operator
\(M\), we define the marginal privileged target as:
\begin{equation}
q_t^\star(\cdot \mid r)
=
\mathbb{E}_{M}
\left[
q\left(\cdot \mid s_t, M(r)\right)
\right].
\label{eq:intro_marginal_target}
\end{equation}

This target has a direct optimization interpretation. Under
the standard teacher-to-student KL objective, matching
\(q_t^\star\) is exactly equivalent, up to a
student-independent constant, to minimizing the expected
distillation loss over randomly sampled privileged views.
Probability-space averaging is therefore not an auxiliary
smoothing heuristic; it is the exact target induced by
expected-view distillation. Predictions that vary when
different parts of the reference are hidden contribute
differently across sampled views, whereas predictions
supported across views persist in the marginal target. In
practice, we approximate the expectation using a small Monte
Carlo sample of independently generated partial views. The
central comparison is therefore:
\begin{equation}
\underbrace{
q(\cdot \mid s_t, r)
}_{\text{one complete privileged view}}
\quad \text{versus} \quad
\underbrace{
\mathbb{E}_{M}
\left[
q(\cdot \mid s_t, M(r))
\right]
}_{\text{marginalized privileged target}}.
\label{eq:central_comparison}
\end{equation}

MCPD should not be interpreted as distillation from a single
incomplete reference. A single partial view remains an
arbitrary realization and may itself induce an unstable
target. As shown in Figure~\ref{fig:reference_sensitivity}, MCPD addresses this issue by treating privileged context as a
distribution over partial views and distilling the resulting
marginal target.
Our main contributions are as follows:
\begin{itemize}
    \item We identify \emph{reference-induced target
    instability} as a measurable limitation of privileged
    on-policy self-distillation.

    \item We introduce MCPD, which replaces a single
    complete-reference target with a teacher distribution
    marginalized over partial privileged views, and derive
    this target directly from expected-view distillation.

    \item Experiments and ablations on mathematical and
    scientific reasoning benchmarks show that MCPD produces
    more reference-consistent supervision and consistently
    improves over competitive baselines.
\end{itemize}

\section{Related Work}

\textbf{Knowledge Distillation.}
Knowledge distillation transfers capabilities from a teacher model to a student through predictive distributions, generated outputs, or intermediate representations~\citep{schmidhuber1992learning,hinton2015distilling,wangabkd}. In language-model reasoning, distillation has enabled smaller models to acquire stronger reasoning~\citep{karan2025reasoning,guha2025openthoughts,ko2025distillm,xu2026rlkd}, instruction-following~\citep{gu2024minillm,ko2024distillm,li-etal-2023-symbolic,wang2025distilling}, and problem-solving abilities~\citep{hubotter2026reinforcement}. Rather than supervising only final answers, recent methods also transfer intermediate reasoning trajectories, allowing the student to learn how a solution is derived~\citep{wangself,li2024mode,li-etal-2023-symbolic}. However, most reasoning-distillation methods remain off-policy: supervision is constructed from fixed reference solutions or teacher-generated trajectories that are independent of the student's own reasoning process~\citep{agarwal2024policy,gu2024minillm,kim-rush-2016-sequence}. The student is therefore trained on expert reasoning prefixes rather than the imperfect states it may encounter during its own generation, creating a mismatch between training-time supervision and inference-time trajectories~\citep{song2026survey}.

\textbf{On-Policy Distillation.}
On-policy distillation addresses this mismatch by providing teacher supervision along trajectories sampled from the student policy~\citep{agarwal2024policy,gu2024minillm,ko2024distillm,jinentropy}. The student first generates a reasoning rollout, after which the teacher provides dense token-level feedback at the same intermediate states. This combines the state coverage of on-policy learning with the fine-grained supervision of knowledge distillation, allowing the model to learn directly from the reasoning states it is likely to visit at inference time.

\textbf{Privileged On-Policy Self-Distillation.}
Recent work extends on-policy distillation to self-distillation, where the same underlying model acts as both student and teacher under different conditioning contexts~\citep{wuinteractions,wu2025native,shenfeld2026self,zhao2026self,song2026survey}. Privileged self-distillation further equips the teacher with information unavailable to the student, such as a verified reference solution~\citep{shenfeld2026self,zhao2026self,nguyen2026avsd,ding2026hdpo}. In OPSD, the teacher observes the complete reference and provides a next-token distribution along the student-generated trajectory~\citep{zhao2026self}. Although this produces informative dense supervision, it implicitly treats the distribution induced by a single complete reference as a reliable target. We instead study how this target changes when the privileged reference changes. We identify \emph{reference-induced target instability}, whereby different valid references can induce substantially different teacher distributions for the same student state. MCPD addresses this issue by sampling partial views of the privileged reference and distilling the marginal teacher distribution obtained by averaging their predictions. Recent work has also explored modifying the privileged context used in
OPSD~\citep{tan2026paint,pan2026rlcsd} but using a single privileged view.
The work most closely related to ours, using multi-views, is
AVSD~\citep{nguyen2026avsd}, which combines three fixed, heterogeneous privileged views: the complete reference solution, a partial solution,
and the final answer. AVSD extracts a consensus component across these
views while selectively retaining view-specific residual information.
However, this design does not directly eliminate dependence on the
particular reference realization. Because the complete-reference
prediction remains one of the supervision sources, and reference-specific
residuals may be preserved, the resulting target can still inherit
information tied to the wording, ordering, or reasoning path of the
given solution. Empirically, MCPD achieves higher average
accuracy than AVSD across all backbones (Table~\ref{tab:qwen3_opsd_results}). 

\section{Method}\label{sec:method}
\subsection{On-Policy Privileged Self-Distillation}

Let
\(
\mathcal{D}=\{(x_i,r_i)\}_{i=1}^{N}
\)
be a reasoning dataset, where \(x_i\) is a problem and \(r_i\)
is a verified reference solution. Given a problem \(x\), the
student policy \(\pi_{\theta}\) generates an on-policy reasoning
trajectory, $
    y=(y_1,\ldots,y_T)
    \sim \pi_{\theta}(\cdot\mid x).
$
At token position \(t\), the student state is
$ s_t=(x,y_{<t}),$
with next-token distribution $
    p_{\theta,t}(\cdot)
    =
    \pi_{\theta}(\cdot\mid s_t).
$

A privileged teacher evaluates the same student state while
additionally observing the reference solution. We denote the
fixed, stop-gradient teacher parameters by \(\bar{\theta}\).
The complete-reference teacher distribution is:
\begin{equation}
q_t^{\mathrm{full}}(\cdot \mid r)
=
\pi_{\bar{\theta}}
\left(\cdot \mid x, r, y_{<t}\right).
\end{equation}

Standard privileged on-policy self-distillation
\citep{zhao2026self} minimizes:

\begin{equation}
\mathcal{L}_{\mathrm{full}}
=
\mathbb{E}_{\substack{(x,r)\sim\mathcal{D}\\
y\sim\pi_{\theta}(\cdot\mid x)}}
\left[
\frac{1}{T}
\sum_{t=1}^{T}
\mathrm{KL}
\left(
q^{\mathrm{full}}_t(\cdot\mid r)
\,\Vert\,
p_{\theta,t}
\right)
\right].
\label{eq:full_reference_loss}
\end{equation}
This objective provides dense supervision along the
student's own trajectory. However, the target at every state
is determined by one complete reference solution.
\subsection{Reference-Induced Target Instability} \label{sub:sens}
As illustrated in Figure~\ref{fig:reference_sensitivity}, changing the
valid reference alone can alter the privileged teacher target. Consider
a problem \(x\), a fixed student-generated trajectory
\(y=(y_1,\ldots,y_T)\), and two valid reference solutions \(r^a\) and
\(r^b\). Let \(\mathcal{I}\subseteq\{1,\ldots,T\}\) denote the matched
rollout positions at which reference sensitivity is evaluated. For both
references, we keep the problem, student trajectory, teacher parameters,
and evaluated positions fixed. The full-reference OPSD teacher
distributions at position \(t\in\mathcal{I}\) are:
\begin{equation}
q_t^{a,\mathrm{full}}
=
\pi_{\bar{\theta}}
\left(\cdot \mid x,r^a,y_{<t}\right),
\qquad
q_t^{b,\mathrm{full}}
=
\pi_{\bar{\theta}}
\left(\cdot \mid x,r^b,y_{<t}\right).
\end{equation}
Because only the privileged reference differs, any discrepancy between
these distributions directly measures dependence on the choice of
reference. We quantify this discrepancy using the Jensen--Shannon (JS)
divergence, which provides a symmetric and bounded comparison between
probability distributions~\citep{lin2002divergence,nielsen2019generalization}. 
Specifically, we define the token-level reference sensitivity of OPSD as:
\begin{equation}
S_t^{\mathrm{OPSD}}
=
\mathrm{JS}
\left(
q_t^{a,\mathrm{full}},
q_t^{b,\mathrm{full}}
\right),
\label{eq:opsd_sensitivity}
\end{equation}
where \(q_t^{a,\mathrm{full}}\) and \(q_t^{b,\mathrm{full}}\) denote the
teacher next-token distributions induced by two valid full references
\(r^a\) and \(r^b\), respectively, for the same problem and student state.
The JS divergence is defined as:
\begin{equation}
\mathrm{JS}(p,q)
=
\frac{1}{2}
D_{\mathrm{KL}}\left(p\,\|\,m\right)
+
\frac{1}{2}
D_{\mathrm{KL}}\left(q\,\|\,m\right),
\qquad
m=\frac{p+q}{2}.
\label{eq:js_divergence}
\end{equation}
A value of zero indicates that the teacher target is invariant to the choice
of reference, whereas a larger value indicates greater reference-induced
variation in the supervision provided at the same student state.

The corresponding trajectory-level sensitivity is:
\begin{equation}
S^{\mathrm{OPSD}}
=
\frac{1}{|\mathcal{I}|}
\sum_{t\in\mathcal{I}}
S_t^{\mathrm{OPSD}}.
\end{equation}

For MCPD, we independently sample $K$ partial views for each
reference:
\begin{equation}
M_{j,1},\ldots,M_{j,K}
\overset{\mathrm{i.i.d.}}{\sim}
\mu_\rho(\cdot\mid r^j),
\qquad j\in\{a,b\}.
\end{equation}
Each view induces the teacher distribution:
\begin{equation}
q_t^{j,(k)}
=
\pi_{\bar{\theta}}
\left(
\cdot\mid x,M_{j,k}(r^j),y_{<t}
\right).
\end{equation}
We construct a separate Monte Carlo marginal target for each
reference:
\begin{equation}
\widehat q_{K,t}^{\,j}
=
\frac{1}{K}
\sum_{k=1}^{K}
q_t^{j,(k)},
\qquad j\in\{a,b\}.
\end{equation}

The token-level and trajectory-level reference sensitivities of
MCPD are then:
\begin{equation}
S_t^{\mathrm{MCPD}}
=
\operatorname{JS}
\left(
\widehat q_{K,t}^{\,a},
\widehat q_{K,t}^{\,b}
\right),
\qquad
S^{\mathrm{MCPD}}
=
\frac{1}{|\mathcal{I}|}
\sum_{t\in\mathcal{I}}
S_t^{\mathrm{MCPD}}.
\end{equation}

Finally, for each problem, we measure the reduction in
reference sensitivity as
$
\Delta S
=
S^{\mathrm{OPSD}}
-
S^{\mathrm{MCPD}}.
$
A positive value of $\Delta S$ indicates that MCPD produces
less reference-dependent teacher targets than full-reference
OPSD for the same problem and student trajectory.
Therefore, $S^{\mathrm{OPSD}}$ isolates variation
induced by the choice of privileged reference. A larger value
indicates that two valid references prescribe more different
token-level supervision for the same student reasoning states.
We refer to this dependence as \emph{reference-induced target
instability}. This metric does not assume that all valid reasoning paths
should induce identical predictions. Rather, it measures the
extent to which the supervision at fixed student states depends
on the particular valid reference supplied to the teacher. Implementation details are provided in Appendix. 

\subsection{Privileged Context as a Distribution}

Rather than treating the complete reference as a single
deterministic privileged context, we define a distribution over
its partial views. Let $M\sim\mu_{\rho}(\cdot\mid r)$
denote a random view operator, where \(\rho\) controls the
fraction of the reference that is hidden. A sampled view
\(M(r)\) reveals only part of the reference while preserving
the ordering of the visible content.

For a fixed student state \(s_t\), each view induces a teacher
predictive distribution:
\begin{equation}
q_t^{M}(\cdot \mid r)
=
\pi_{\bar{\theta}}
\left(\cdot \mid x,M(r),y_{<t}\right).
\end{equation}

All views use the same underlying teacher and the same
student state; only the privileged conditioning context
changes. Thus, MCPD uses one teacher under multiple
views.
We define the marginal predictive target as:
\begin{equation}
q^\star_t(\cdot\mid r)
=
\mathbb{E}_{M\sim\mu_{\rho}(\cdot\mid r)}
\left[
q^M_t(\cdot\mid r)
\right].
\label{eq:marginal_target}
\end{equation}
The expectation is taken directly in probability space.
Because every \(q^M_t\) is a categorical distribution,
\(q^\star_t\) is also a valid categorical distribution. This construction marginalizes over partial observations of
the available reference. 
\subsection{Expected-View Distillation}

We show that the probability-space average in
Eq.~\eqref{eq:marginal_target} is the exact target induced by
distillation over random partial views, rather than a heuristic
aggregation rule. For compactness, we suppress the explicit conditioning
on \(r\) and write \(\mathbb{E}_{M}\) for
\(\mathbb{E}_{M\sim\mu_{\rho}(\cdot\mid r)}\).
For a fixed student state \(s_t\) and reference \(r\), consider:
\begin{equation}
\mathcal{J}_t(p)
=
\mathbb{E}_{M}
\left[
\mathrm{KL}
\left(
q_t^M
\,\Vert\,
p
\right)
\right],
\label{eq:expected_view_objective}
\end{equation}
where \(p\) is any categorical student distribution.

\noindent\textbf{Proposition 1.}
Let \(q_t^\star\) be defined as in
Eq.~\eqref{eq:marginal_target}. Then,
\begin{equation}
\mathcal{J}_t(p)
=
\mathbb{E}_{M}
\left[
\mathrm{KL}
\left(
q_t^M
\,\Vert\,
q_t^\star
\right)
\right]
+
\mathrm{KL}
\left(
q_t^\star
\,\Vert\,
p
\right).
\label{eq:kl_decomposition}
\end{equation}

Consequently,
\begin{equation}
\operatorname{arg\,min}_{p}\,
\mathcal{J}_t(p)
=
q_t^\star.
\label{eq:expected_view_optimum}
\end{equation}

\noindent\textit{Proof.}
Expanding the expected KL divergence gives
\begin{align}
\mathcal{J}_t(p)
&=
\mathbb{E}_{M}
\left[
\sum_{v\in\mathcal{V}}
q_t^M(v)
\log
\frac{q_t^M(v)}{p(v)}
\right]
\nonumber\\
&=
\mathbb{E}_{M}
\left[
\sum_{v\in\mathcal{V}}
q_t^M(v)
\log
\frac{q_t^M(v)}{q_t^\star(v)}
\right]
\nonumber\\
&\quad+
\sum_{v\in\mathcal{V}}
q_t^\star(v)
\log
\frac{q_t^\star(v)}{p(v)}.
\end{align}
This yields Eq.~\eqref{eq:kl_decomposition}. The first term is
independent of \(p\), while the second is minimized when
\(p=q_t^\star\).
\hfill\(\square\)

Thus, matching the view-marginalized target is equivalent, up to a
student-independent term, to minimizing the expected forward-KL loss
over random partial views. The first term in
Eq.~\eqref{eq:kl_decomposition} captures disagreement among the
view-conditioned teacher predictions, whereas the second measures the
student's divergence from their view-marginalized target.

\subsection{Monte Carlo Approximation}

Computing \(q^\star_t\) exactly would require evaluating all
possible partial views. We approximate it using \(K\)
independently sampled views:
$
M_1,\ldots,M_K
\overset{\mathrm{i.i.d.}}{\sim}
\mu_{\rho}(\cdot\mid r).
$
The Monte Carlo target is:
$
\widehat{q}_{K,t}(\cdot\mid r)
=
\frac{1}{K}
\sum_{k=1}^{K}
q^{M_k}_t(\cdot\mid r).
\label{eq:mc_target}
$
For a fixed state and reference, this estimator is unbiased:
\begin{equation}
\mathbb{E}
\left[
\widehat{q}_{K,t}(\cdot\mid r)
\right]
=
q^\star_t(\cdot\mid r).
\end{equation}
Its covariance satisfies:
\begin{equation}
\mathrm{Cov}
\left[
\widehat{q}_{K,t}(\cdot\mid r)
\right]
=
\frac{1}{K}
\mathrm{Cov}
\left[
q^M_t(\cdot\mid r)
\right].
\label{eq:mc_covariance}
\end{equation}
Thus, the variability caused by sampling a finite set of
views decreases proportionally to \(1/K\). This guarantee concerns Monte Carlo estimation for a fixed
reference. It does not imply that:
\begin{equation}
q^\star_t(\cdot\mid r^a)
=
q^\star_t(\cdot\mid r^b)
\end{equation}
for two distinct valid references. Reduced cross-reference
sensitivity is therefore an empirical property of MCPD,
rather than a direct theoretical consequence of the
\(1/K\) variance reduction.

\subsection{Monte Carlo Privileged Distillation}

MCPD trains the student against the Monte Carlo marginal
target:
\begin{equation}
\mathcal{L}_{\mathrm{MCPD}}
=
\mathbb{E}_{\substack{(x,r)\sim\mathcal{D}\\
y\sim\pi_{\theta}(\cdot\mid x)\\
M_{1:K}\sim\mu_{\rho}(\cdot\mid r)}}
\left[
\frac{1}{T}
\sum_{t=1}^{T}
\mathrm{KL}
\left(
\operatorname{sg}
\left[
\widehat{q}_{K,t}(\cdot\mid r)
\right]
\,\Vert\,
p_{\theta,t}
\right)
\right].
\label{eq:mcppd_loss}
\end{equation}

For each training example, the procedure consists of four
steps. First, the student generates one on-policy trajectory
from the problem alone. Second, we sample \(K\) independent
partial views of the same reference. Third, the teacher
evaluates the same student states under every privileged
view. Finally, the resulting next-token distributions are
averaged at each position, and the student is updated toward
the marginal target. All teacher evaluations share the same problem and
student-generated prefix. Consequently, variation among the
teacher distributions arises only from changes in privileged
context, rather than from different student trajectories.

In our implementation, each partial view is generated as
\begin{equation}
M_k(r)
=
\operatorname{Mask}(r;\rho,\xi_k),
\label{eq:masking_operator}
\end{equation}
where \(\xi_k\) denotes an independently sampled masking
pattern. Each view hides a fraction \(\rho\)
of the reference while preserving the order of the visible
content. The complete reference, partial views, and masking
patterns are used only by the stop-gradient teacher and are
never observed by the student. The complete algorithm is provided in Appendix.
MCPD requires one student rollout and \(K\) teacher evaluations per
training example. Since all views share the same student states, the
teacher evaluations can be batched across views. Increasing \(K\)
therefore trades additional teacher-side computation for a lower-variance
approximation of the marginal predictive target. In practice, with
\(K=4\), MCPD requires only modest additional wall-clock time:
on \texttt{deepseek-ai/DeepSeek-R1-Distill-Qwen-7B}, it trains in
1h 36m, only 10 minutes longer than OPSD and slightly faster than AVSD
under the same 100-step training budget. Under its 500-step,
8-rollout configuration, GRPO requires 12h 30m. Full computational-cost results are provided in
Appendix Table~2.

\section{Experiments}
\label{sec:math_experiments}

\subsection{Mathematical Reasoning}
\begin{table}[t]
\centering
\small
\setlength{\tabcolsep}{6pt}
\renewcommand{\arraystretch}{1.08}
\begin{tabular}{l|cccc}
\toprule
\textbf{Method} & \textbf{AIME24} & \textbf{AIME25} & \textbf{HMMT25} & \textbf{AVG} \\
\midrule

\multicolumn{5}{l}{\textit{Qwen3-14B}} \\
\quad Base (Instruct) & 80.8 & 68.9 & 50.0 & 66.6\\
\quad + SFT          & 79.7 & 73.9 & 49.7  & 67.7 \\
\quad + GRPO         & 81.9 & 73.6 & \textbf{53.0} & 69.5 \\
\quad + OPSD        &81.1 & 73.9 & 52.2 & 69.1   \\
\quad + AVSD        &80.6 & 73.6 & 50.6  &  68.3 \\
\quad + MCPD (ours)        &\textbf{82.2} & \textbf{75.0} & 51.7 & \textbf{69.6}   \\

\midrule

\multicolumn{5}{l}{\textit{Qwen3-8B}} \\
\quad Base (Instruct) & 75.8 & 65.6 & 43.9 & 61.8 \\
\quad + SFT          & 72.3 & 64.2 & 42.9 & 59.8 \\
\quad + GRPO         & 76.4 & 68.9 & \textbf{46.7} & 64.0 \\
\quad + OPSD         & 77.8 & 70.8 & 45.8 & 64.8 \\
\quad + AVSD        & 77.5 & 69.4 & 45.3 &  64.1 \\
\quad + MCPD (ours)         & \textbf{78.9} & \textbf{72.8} &\textbf{46.7}  &\textbf{ 66.1} \\

\midrule

\multicolumn{5}{l}{\textit{Qwen3-4B}} \\
\quad Base (Instruct) & 74.9 & 66.4 & 42.2 & 61.2 \\
\quad + SFT          & 70.2 & 62.3 & 43.4 & 58.6 \\
\quad + GRPO         & 75.6 & 68.1 & 44.4 & 62.7 \\
\quad + OPSD         & \textbf{76.4} & 68.3 & 46.1 & 63.6 \\
\quad + AVSD        &74.4 & 65.0 & 47.2 &  62.2 \\
\quad + MCPD (ours)         & 75.6 & \textbf{70.6 }& \textbf{48.1} &\textbf{ 64.8}  \\
\midrule

\multicolumn{5}{l}{\textit{DeepSeek-R1-Distill-Qwen-7B }} \\

\quad + SFT         & 54.2 & 38.6 &  23.9&38.9  \\
\quad + OPSD         & 54.4 &41.7  &  24.1&40.1  \\
\quad + AVSD        & 54.2& 40.3 & 25.0 & 39.8\\
\quad + MCPD (ours)        & \textbf{55.0} &\textbf{43.1} &\textbf{25.6} & \textbf{41.2} \\
\bottomrule
\end{tabular}
\caption{Avg@12 accuracy on math benchmarks across four backbones. Best results
are in bold. AVG refers to average.}
\label{tab:qwen3_opsd_results}
\end{table}

\paragraph{Models and data.}
We follow the mathematical-reasoning setup of OPSD~\citep{zhao2026self}.
We experiment with three instruction-tuned Qwen3 backbones~\citep{yang2025qwen3}: \texttt{Qwen3-4B}, \texttt{Qwen3-8B}, and \texttt{Qwen3-14B}, as well as \texttt{DeepSeek-R1-Distill-Qwen-7B}~\citep{guo2025deepseek}.
All post-training methods use the exact 17K-example mathematical-reasoning subset of OpenThoughts used by OPSD~\citep{zhao2026self,guha2025openthoughts}, containing problems paired with expert reference solutions.
We report our main results on three competition-level
benchmarks: AIME 2024 and AIME 2025~\citep{maa2025aime},
and HMMT 2025~\citep{balunovic2026matharena}.
AMO-Bench~\citep{liu2026amo} is additionally used in
the masking-ratio and number-of-views ablations. Full implementation details are provided in Appendix. 
\paragraph{Baselines.}
We compare MCPD with the original instruction-tuned models and four post-training baselines:
\textsc{SFT}~\citep{weifinetuned,longpre2023flan},
\textsc{GRPO}~\citep{shao2024deepseekmath},
\textsc{OPSD}~\citep{zhao2026self}, and
\textsc{AVSD}~\citep{nguyen2026avsd}, which conditions the teacher on three privileged views, the complete solution, a partial solution, and the final answer, and combines their predictions using its proposed gating mechanism; we use a partial-solution ratio of $0.5$ as in \citet{nguyen2026avsd}.
Our method, \textsc{MCPD}, constructs $K=4$ partial views of the reference solution for each training example.
For each view, we randomly mask a fixed fraction of the token positions within the reference-solution span.
The masking pattern is sampled independently for each view and resampled in every training batch.
We mask $30\%$ of the reference tokens for the Qwen3 models and $40\%$ for \texttt{DeepSeek-R1-Distill-Qwen-7B}. 
The same frozen teacher evaluates the student states under all views, 
and the resulting next-token distributions are averaged in probability space to form the Monte Carlo marginal target. The evaluation setup is detailed in Appendix.

\paragraph{Results.}
Table~\ref{tab:qwen3_opsd_results} shows that our MCPD achieves the highest average accuracy among all competing methods for each of the four backbones.
The gains are particularly strong on AIME25: compared with OPSD, MCPD improves accuracy by $2.0$ percentage points on \texttt{Qwen3-8B} and by $2.3$ points on \texttt{Qwen3-4B}.
Compared with AVSD, the improvement reaches $5.6$ points on AIME25 with \texttt{Qwen3-4B}.
MCPD also surpasses GRPO in average accuracy across all Qwen3 models.
Overall, these results demonstrate that MCPD improves downstream reasoning performance.

\subsection{Science Reasoning} \label{subsec:SR}

\paragraph{Models and data.}
We evaluate our method on the SciKnowEval L3 multiple-choice scientific reasoning benchmark~\citep{feng2024sciknoweval}, covering biology, chemistry, physics, and materials science.
We use the same per-domain SciKnowEval splits as in~\citet{hubotter2026reinforcement}, comprising $3{,}868$ training examples and $429$ held-out test examples.
The training set contains $450$ biology, $1{,}890$ chemistry, $687$ physics, and $841$ materials-science examples.
All experiments use \texttt{Qwen3-8B} as the backbone and follow the matched OPSD training configuration of~\citet{zhao2026self}, with a maximum student completion length of $2048$ tokens.
For MCPD, we use $K=4$ independently sampled partial reference views with a masking ratio of $\rho=10\%$. The evaluation setup is detailed in the Appendix.
\paragraph{Results.}
Table~\ref{tab:science_tooluse_results} shows that MCPD achieves the highest overall accuracy on SciKnowEval, outperforming both OPSD and AVSD on average.
Compared with OPSD, MCPD improves the average score from $49.7$ to $50.9$, a gain of $1.2$ percentage points, with particularly strong improvements of $2.4$ points in Biology and $1.8$ points in Physics.
Relative to AVSD, MCPD achieves a $3.0$-point gain in Physics.
\begin{table}[h]
\centering
\small
\setlength{\tabcolsep}{6pt}
\renewcommand{\arraystretch}{1.08}
\begin{tabular}{lcccccc}
\toprule
\textbf{Method} & \textbf{Chem} & \textbf{Physics} & \textbf{Biology} & \textbf{Materials}& AVG \\
\midrule

\textbf{Qwen3-8B} & 41.2 & 59.2 & 30.8 & 58.9 &47.5 \\
\quad + OPSD & 44.1 & 60.4&  35.6& 58.6 & 49.7\\
\quad + AVSD & \textbf{44.3} & 59.2& 37.2 & \textbf{60.6} & 50.3\\
\quad + MCPD & \textbf{44.3}& \textbf{62.2} & \textbf{38.0}&  59.0 & \textbf{50.9}\\

\bottomrule
\end{tabular}
\caption{Avg@16 across scientific reasoning tasks using Qwen3-8B. Best results are in bold. AVG refers to average.}
\label{tab:science_tooluse_results}
\end{table}

\paragraph{Out-of-Domain Science Reasoning}

We further examine whether the reasoning capabilities acquired through mathematical training transfer to an unseen domain.
Specifically, we evaluate the \texttt{Qwen3-8B} models trained on mathematical reasoning with OPSD, AVSD, and MCPD on the held-out SciKnowEval L3 test sets, following the evaluation protocol in the experiment above.
As shown in Table~\ref{tab:ood_science_results}, MCPD achieves the highest average score.
It obtains the strongest performance in Chemistry, Physics, and Materials, while matching AVSD in Biology.
Consequently, MCPD improves in-domain mathematical-reasoning performance and transfers effectively to out-of-domain scientific reasoning.

\begin{table}[t]
\centering
\small
\begin{tabular}{lccccc}
\toprule
\textbf{Method} & \textbf{Chem} & \textbf{Physics} & \textbf{Biology} & \textbf{Materials} & \textbf{AVG} \\
\midrule
OPSD & 41.2 & 59.7 & 31.1 & 58.7 & 47.7 \\
AVSD & 41.7 &  59.2&  \textbf{32.9}& 59.4 &  48.3\\
MCPD & \textbf{42.4} & \textbf{60.5} & \textbf{32.9} & \textbf{60.3} & \textbf{49.0} \\
\bottomrule
\end{tabular}
\caption{Avg@16 comparison of math-trained Qwen3-8B models with OPSD, AVSD, and our MCPD, on SciKnowEval scientific-reasoning domains. AVG refers to average.}
\vspace{- 2 em}
\label{tab:ood_science_results}
\end{table}

\subsection{Ablation Studies}
\begin{figure*}[t]
    \centering

    \begin{subfigure}[t]{0.64\textwidth}
        \centering
        \includegraphics[width=\linewidth]{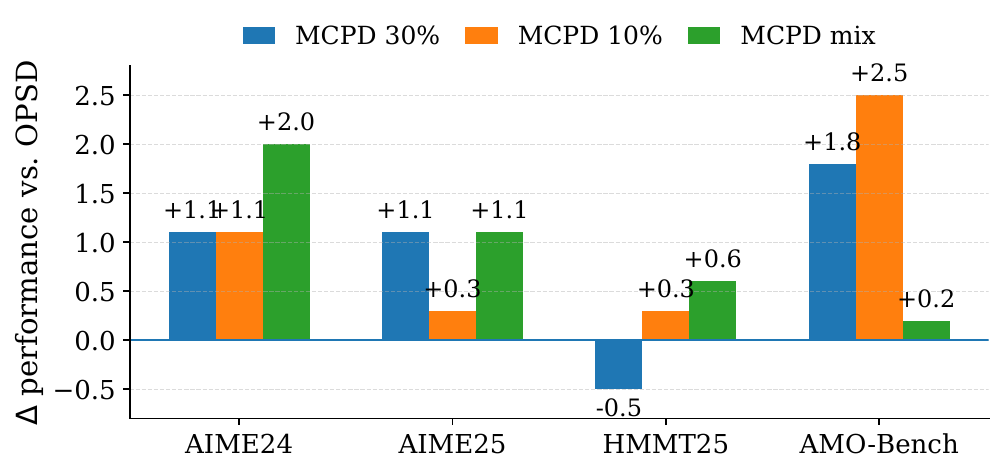}
    \end{subfigure}
    \hfill
    \begin{subfigure}[t]{0.32\textwidth}
        \centering
        \includegraphics[width=\linewidth]{
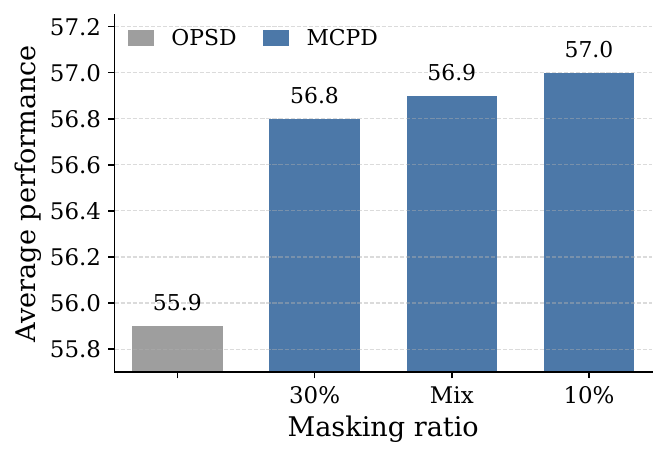}
    \end{subfigure}
\caption{
Effect of reference-solution masking on Qwen3-14B.
\textbf{Left:} Performance changes relative to OPSD across
individual benchmarks.
\textbf{Right:} Average performance for OPSD and each masking
configuration across AIME24, AIME25, HMMT25, and AMO-Bench.
Mix denotes mixed masking ratios of $0\%$, $5\%$, $10\%$, and
$30\%$.
}
    \label{fig:masking_ratio}
\end{figure*}

\begin{figure}[t]
    \centering
    \includegraphics[width=0.8\linewidth]{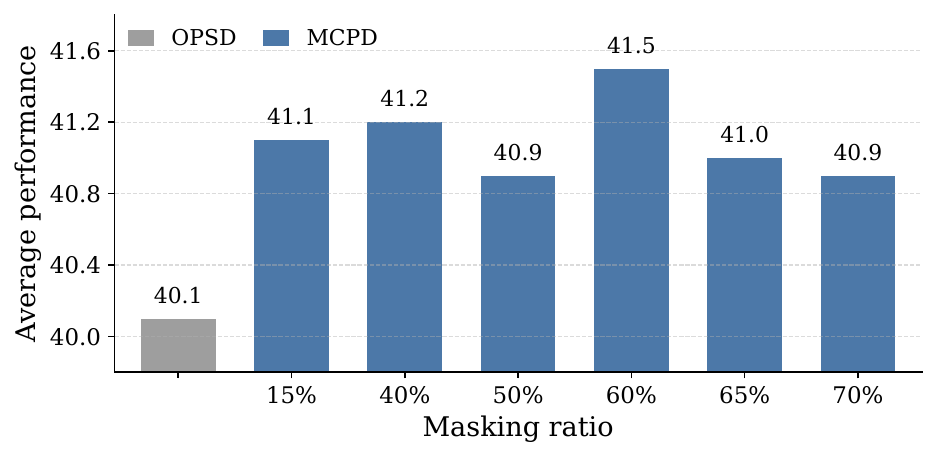}
\caption{
MCPD with different masking ratios on
DeepSeek-R1-Distill-Qwen-7B, with the number of views fixed at
$K=4$. Full results are provided in the Appendix.
}
    \label{fig:deepseekmaskingratio}
\end{figure}

\paragraph{Different masking ratios.}
Figure~\ref{fig:masking_ratio} examines the sensitivity of MCPD to the
masking ratio across two model families. The number of views is fixed to $K=4$.
On \texttt{Qwen3-14B}, all evaluated masking configurations improve the
average score over OPSD, increasing it from $55.9$ to
$56.8$--$57.0$.
We observe similar robustness on
\texttt{DeepSeek-R1-Distill-Qwen-7B} (Figure~\ref{fig:deepseekmaskingratio}). All evaluated masking ratios
achieve higher average accuracy than both OPSD and AVSD. Across all masking ratios, MCPD obtains average scores between $40.9$ and $41.5$, compared with
$40.1$ for OPSD and $39.8$ for AVSD.

\begin{figure}[t]
    \centering
    \includegraphics[width=0.52\linewidth]
    {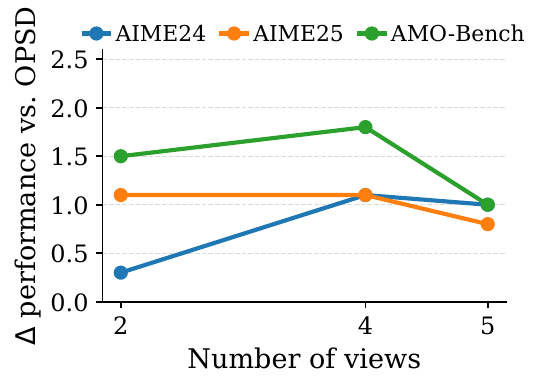}
    \hfill
    \includegraphics[width=0.44\linewidth]
    {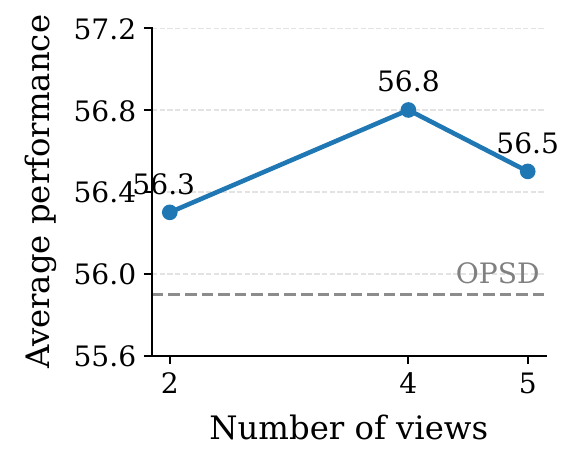}

    \caption{
    Effect of the number of views. \textbf{Left}: Performance changes relative to OPSD on three benchmarks. \textbf{Right}: Average performance across AIME24, AIME25,
HMMT25, and AMO-Bench. The masking ratio is fixed at $\rho=30\%$.
    }
    \vspace{-2 em}\label{fig:view_ablation}
\end{figure}

\begin{figure}[t]
    \centering
    \includegraphics[width=0.6\linewidth]{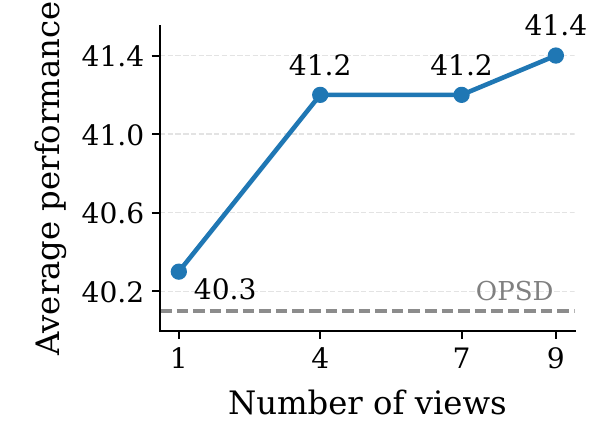}
    \caption{MCPD with different numbers of views on DeepSeek-R1-Distill-Qwen-7B with the masking ratio fixed at $\rho=40\%$. Full results table in Appendix.}
   
    \label{fig:deepseeknumberofviews}
\end{figure}

\paragraph{Different numbers of views.}
We study the effect of the number of sampled privileged views, $K$, on
\texttt{Qwen3-14B} (Figure~\ref{fig:view_ablation}) and
\texttt{DeepSeek-R1-Distill-Qwen-7B}
(Figure~\ref{fig:deepseeknumberofviews}). Across the evaluated values of $K$, MCPD exceeds
OPSD and remains stronger than AVSD, demonstrating robustness
to the number of sampled views.
The DeepSeek ablation further highlights the importance of aggregating
multiple views: using a single randomly masked view achieves an average
performance of $40.3$, whereas averaging $K=4$ views increases it to
$41.2$.
This gap shows that one partial view can produce an unreliable target
that depends on a particular masking realization.
Monte Carlo averaging instead integrates predictions across several
partial views, yielding a more reliable distillation target.

\paragraph{Sampled-token distillation.}
We evaluate whether MCPD remains effective when distillation is restricted to the sampled student token rather than the full vocabulary. Table~\ref{tab:sampledtoken}
shows that MCPD achieves the strongest overall performance, improving over OPSD by $4.6$ percentage points on AIME25 and $9.4$ points on HMMT25.
Compared with AVSD, MCPD matches its AIME25 accuracy while improving HMMT25 by $3.4$ points.
These results indicate that MCPD remains effective under sampled-token supervision.
\begin{table}[ht]
\centering
\small
\begin{tabular}{l|cc}
\toprule
\textbf{Method } & \textbf{AIME25} & \textbf{HMMT25} \\
\midrule
OPSD w/ Sampled-token distillation 
& 82.1 & 57.3 \\
AVSD w/ Sampled-token distillation 
& 86.7&  63.3 \\
MCPD w/ Sampled-token distillation 
&\textbf{86.7} & \textbf{66.7} \\
\bottomrule
\end{tabular}
\caption{Sampled-token distillation results on Qwen3-4B
with a maximum student completion length of 2,048 tokens.
We report $\mathrm{Pass@8}$ accuracy on AIME25 and HMMT25 as in~\citet{zhao2026self}.}
\label{tab:sampledtoken}
\vspace{-2 em}
\end{table}
\section{Conclusion}
In this work, we identified reference-induced target instability as an important limitation of privileged OPSD: even when the problem and student state are fixed, replacing one valid reference solution with another can induce substantially different teacher targets. To address this issue, we introduced Monte Carlo Privileged Distillation (MCPD), which treats privileged context as a distribution over partial reference views and distills the resulting marginal teacher distribution. We showed that averaging the view-conditioned predictions in probability space is the exact target induced by minimizing the expected teacher-to-student forward-KL objective over privileged views. Empirically, MCPD produces more reference-consistent supervision and achieves the strongest average performance across four mathematical-reasoning backbones, while also improving scientific reasoning and out-of-domain transfer from mathematics to science. Ablations further show that the method is robust across masking ratios, benefits from multiple partial views, and remains effective under sampled-token distillation. Therefore, privileged supervision should not be tied to a
single complete realization of a solution. Marginalizing over partial views
provides a simple and effective way to preserve useful reference information
while reducing sensitivity to reference-specific artifacts and
reasoning paths. Future work includes extending our approach MCPD beyond mathematical and scientific reasoning, e.g., multi-modal reasoning settings.

\newpage
\bibliography{aaai2027}
\newpage
\appendix
\section{Target Entropy Analysis}

We examine whether MCPD's lower reference sensitivity can be
explained by producing more diffuse teacher targets. We use
270 problems, together with the
same reference pairs, student trajectories, and matched token
positions used in the reference-sensitivity analysis.

For method $m \in \{\mathrm{OPSD}, \mathrm{MCPD}\}$,
reference $j \in \{a,b\}$, and matched position
$t \in \mathcal{I}_i$, let $q_{i,t}^{m,j}$ denote the corresponding
full-vocabulary target distribution. For OPSD, this is the
teacher distribution conditioned on the complete reference. For
MCPD, it is the probability-space average of the $K=4$
view-conditioned teacher distributions. We compute the Shannon entropy of each target as:
\begin{equation}
H\!\left(q_{i,t}^{m,j}\right)
=
-\sum_{v\in\mathcal{V}}
q_{i,t}^{m,j}(v)
\log q_{i,t}^{m,j}(v),
\end{equation}
where $\mathcal{V}$ is the model vocabulary. We use the natural
logarithm, so entropy is measured in nats.

For each problem $i$, we average the entropy across both references
and all matched student states:
\begin{equation}
H_i^m
=
\frac{1}{2|\mathcal{I}_i|}
\sum_{j\in\{a,b\}}
\sum_{t\in\mathcal{I}_i}
H\!\left(q_{i,t}^{m,j}\right).
\end{equation}

We report the mean problem-level entropy and the paired rate, defined
as the percentage of problems for which
$H_i^{\mathrm{MCPD}} < H_i^{\mathrm{OPSD}}$.
As shown in Table~\ref{tab:entropy_analysis}, MCPD has comparable,
and slightly lower, mean entropy than OPSD and produces lower entropy
on $75.6\%$ of the evaluated problems. Thus, MCPD's reduction in
reference sensitivity is not explained by flattening the target
distribution.

\begin{table}[t]
\centering
\caption{Full-vocabulary target entropy at the matched student states
used for reference-sensitivity evaluation. The paired rate is the
percentage of problems on which MCPD produces lower problem-level
entropy than OPSD.}
\label{tab:entropy_analysis}
\begin{tabular}{lcc}
\toprule
\textbf{Method} & \textbf{Mean entropy} & \textbf{Paired rate} \\
\midrule
OPSD & 0.34 & -- \\
MCPD & 0.33 & 75.6\% \\
\bottomrule
\end{tabular}
\end{table}
\section{Reference-Sensitivity Implementation Details}

We evaluate reference sensitivity on 500 problems sampled from the
training split of DeepMath-103K~\citep{he2026deepmath} using seed 42
and Qwen3-8B~\citep{yang2025qwen3}. The same underlying checkpoint
is used to generate the unprivileged student rollout and as the
frozen privileged teacher. DeepMath-103K provides three verified
reference solutions in the fields
\texttt{r1\_solution\_1},
\texttt{r1\_solution\_2}, and
\texttt{r1\_solution\_3}. For each selected problem $x_i$, we use
the first two verified solutions, denoted $r_i^a$ and $r_i^b$.

For each problem, we first generate one unprivileged student rollout
$y_i$ from the problem alone using the Qwen chat template with
\texttt{enable\_thinking=False}. Rollouts are sampled with
temperature $0.6$, top-$p=0.95$, and a maximum of 1,024 new tokens.
The resulting rollout is then held fixed and reused for both
references and for both OPSD and MCPD. We evaluate teacher
distributions at 20 normalized prefix positions uniformly spaced
between 10\% and 90\% of the rollout length. These positions form
the matched evaluation set $\mathcal{I}_i$.

For all target evaluations, the frozen teacher uses the Qwen chat
template with \texttt{enable\_thinking=True}. This configuration is
used for both OPSD and MCPD. The teacher does not generate a separate
trajectory; it only returns next-token distributions at matched
prefixes of the fixed student rollout. For OPSD, the teacher conditions on the complete reference solution.
For reference $j\in\{a,b\}$ and position $t\in\mathcal{I}_i$, the
full-reference teacher distribution is:
\begin{equation}
q_{i,t}^{j,\mathrm{full}}
=
\pi_{\bar{\theta}}
\left(
\cdot \mid x_i,r_i^j,y_{i,<t}
\right),
\end{equation}
where $\bar{\theta}$ denotes the frozen initial checkpoint.

For MCPD, we independently sample $K=4$ partial views of each
reference by masking 30\% of its reference-token positions. The
masking pattern is sampled independently for each view, and the
ordering of the unmasked tokens is preserved. 
Let $M_{s,k}$ denote the $k$-th masking
operator in mask bank $s$. The resulting Monte Carlo marginal target is:

\begin{equation}
\widehat q_{i,t,s}^{j}
=
\frac{1}{K}
\sum_{k=1}^{K}
\pi_{\bar{\theta}}
\left(
\cdot \mid x_i,M_{s,k}(r_i^j),y_{i,<t}
\right),
\qquad K=4.
\end{equation}

To reduce Monte Carlo noise in the sensitivity estimate, we repeat
the masking procedure using five reproducible mask banks generated
with seeds:
\[
\mathcal{S}=\{42,43,44,45,46\}.
\]
For each mask bank, we first average the four view-conditioned
teacher distributions in probability space and then compute the
reference sensitivity between the two resulting marginal targets.
\paragraph{Reference-sensitivity metric.}
We measure the discrepancy between the teacher distributions induced
by two valid references using the Jensen--Shannon (JS) divergence~\citep{lin2002divergence}.
For two categorical distributions $p$ and $q$, let
$m=\frac{1}{2}(p+q)$. The JS divergence is defined as:
\begin{equation}
\operatorname{JS}(p,q)
=
\frac{1}{2}D_{\mathrm{KL}}(p\|m)
+
\frac{1}{2}D_{\mathrm{KL}}(q\|m).
\end{equation}
We use JS divergence because it is symmetric and bounded, and therefore
does not privilege either reference as the comparison direction. It
also remains finite when the two distributions assign probability mass
to different vocabulary tokens.

Using the natural logarithm, its range is:
\begin{equation}
0
\leq
\operatorname{JS}(p,q)
\leq
\log 2.
\end{equation}
A value of zero indicates identical teacher distributions, whereas
larger values indicate greater dependence on the selected reference.
The maximum value $\log 2 \approx 0.693$ is attained when the two
distributions have disjoint support.
For each problem, the reference sensitivity of OPSD is:
\begin{equation}
S_i^{\mathrm{OPSD}}
=
\frac{1}{|\mathcal{I}_i|}
\sum_{t\in\mathcal{I}_i}
\operatorname{JS}
\left(
q_{i,t}^{a,\mathrm{full}},
q_{i,t}^{b,\mathrm{full}}
\right).
\end{equation}

The corresponding reference sensitivity of MCPD is:
\begin{equation}
S_i^{\mathrm{MCPD}}
=
\frac{1}{|\mathcal{I}_i||\mathcal{S}|}
\sum_{t\in\mathcal{I}_i}
\sum_{s\in\mathcal{S}}
\operatorname{JS}
\left(
\widehat q_{i,t,s}^{a},
\widehat q_{i,t,s}^{b}
\right).
\end{equation}

The per-problem sensitivity reduction plotted in Figure~1 of the
main paper is:
\begin{equation}
\Delta S_i
=
S_i^{\mathrm{OPSD}}
-
S_i^{\mathrm{MCPD}}.
\end{equation}
A positive value indicates that MCPD produces a less
reference-dependent target than full-reference OPSD for the same
problem, student trajectory, reference pair, and matched student
states. MCPD has lower reference sensitivity than OPSD on 80.4\% of
the 500 evaluated problems.
\section{Computational Cost}

We compare the training cost of all methods using
\texttt{deepseek-ai/DeepSeek-R1-Distill-Qwen-7B}. All timing
experiments are conducted on the same hardware configuration of four
NVIDIA A100 80GB GPUs. OPSD, MCPD, and AVSD are each trained for
100 optimization steps, whereas GRPO is trained for 500 optimization
steps with eight sampled rollouts per prompt, following the
configurations used in our experiments. MCPD uses $K=4$ partial
reference views per training example. Table~\ref{tab:training_cost}
reports the wall-clock time of one complete training run for each
method.

\begin{table}[t]
\centering
\caption{Wall-clock training time on
\texttt{DeepSeek-R1-Distill-Qwen-7B} using four NVIDIA A100 80GB
GPUs. Times correspond to the method-specific training configurations
used in our experiments.}
\label{tab:training_cost}
\begin{tabular}{lc}
\toprule
\textbf{Method} & \textbf{Training time} \\
\midrule
OPSD & 1h 26m \\
MCPD (ours) & 1h 36m \\
AVSD & 1h 40m \\
GRPO & 12h 30m \\
\bottomrule
\end{tabular}
\end{table}

MCPD requires 10 additional minutes relative to OPSD, corresponding
to approximately 11.6\% additional wall-clock time, while remaining
comparable in cost to AVSD. Under its 500-step and eight-rollout
configuration, GRPO requires approximately $7.8\times$ the wall-clock
training time of MCPD. 
\section{Experiments}

\subsection{Mathematical Reasoning}

\paragraph{Training configuration.}
We follow the mathematical-reasoning setup of
OPSD~\citep{zhao2026self}, in which the same underlying language model acts as an on-policy student and a frozen privileged teacher under different conditioning contexts. The student conditions only on the problem and generates an on-policy reasoning trajectory. The teacher evaluates the same student states while additionally conditioning on privileged reference information and provides dense token-level
supervision. 
Following~\citet{zhao2026self}, all post-training methods use
LoRA~\citep{hu2022lora} with rank 64 and scaling factor 128. The
adapters are applied to the \texttt{q\_proj}, \texttt{k\_proj},
\texttt{v\_proj}, \texttt{o\_proj}, \texttt{gate\_proj},
\texttt{up\_proj}, and \texttt{down\_proj} modules. Unless otherwise
specified, we optimize the LoRA parameters using AdamW with a learning
rate of $5\times10^{-6}$, an effective batch size of 32, and a maximum
gradient norm of 0.1. OPSD, AVSD, SFT, and MCPD are each trained for 100
optimization steps, with checkpoints saved every 25 steps.

Student rollouts are sampled with temperature $1.1$, top-$p=0.95$,
top-$k=20$, and a maximum completion length of 1,024 tokens. The
maximum context length is 20,000 tokens. Student prompts are
constructed with thinking disabled, whereas privileged teacher prompts
are constructed with thinking enabled whenever the tokenizer supports
the \texttt{enable\_thinking} argument.

\paragraph{Teacher and optimization objective.}
The teacher is fixed to the initial model checkpoint throughout
training. Only the student's LoRA parameters are updated; the teacher
parameters receive no gradients and do not incorporate the updated
student adapters. OPSD, AVSD, and MCPD minimize the full-vocabulary
forward KL divergence,
\begin{equation}
D_{\mathrm{KL}}\!\left(p_T \,\|\, p_S\right),
\end{equation}
between the privileged teacher and student next-token distributions at
each valid position of the student-generated rollout.

\paragraph{MCPD configuration.}
MCPD uses $K=4$ independently sampled partial reference views per
training example. For the Qwen3 models, each view masks 30\% of the
reference-solution token positions. For
\texttt{DeepSeek-R1-Distill-Qwen-7B}, each view masks 40\% of the
reference-token positions. Selected tokens are replaced with the
tokenizer's \texttt{[MASK]} token, while all unmasked tokens retain
their original order. Masking patterns are independently sampled for
the four views and resampled in every training batch. The frozen
teacher evaluates the same student states under all four views, and
the resulting next-token distributions are averaged in probability
space before computing the distillation loss.

\paragraph{Baselines.}
For AVSD, we use the complete-solution, partial-solution, and
final-answer privileged views and set the partial-solution ratio to
0.5, following~\citet{nguyen2026avsd}. The SFT baseline uses the same
LoRA configuration, optimizer, effective batch size, and numerical
precision as the distillation methods. 
For GRPO, we sample eight responses per prompt with temperature $1.2$,
use binary answer-verification rewards, set the KL coefficient to zero,
use a maximum prompt length of 2,048 tokens, and train for 500
optimization steps.

\subsection{Evaluation Setup}

\paragraph{Evaluation metrics.}
For a benchmark containing $N$ problems, let
$c_{i,k}\in\{0,1\}$ indicate whether the $k$-th independently
sampled response to problem $i$ is correct. We define
\begin{equation}
\operatorname{Avg@K}
=
\frac{1}{N}
\sum_{i=1}^{N}
\frac{1}{K}
\sum_{k=1}^{K}
c_{i,k}.
\end{equation}
Thus, Avg@12 and Avg@16 report the mean correctness across 12 and
16 independently sampled responses per problem, respectively.
Unlike a success-at-least-once metric, Avg@K~\citep{zhao2026self,hubotter2026reinforcement} estimates the expected
accuracy of a single response under stochastic decoding.

\paragraph{Mathematical-reasoning evaluation.}
We follow the evaluation setup of~\citet{zhao2026self}. For each
problem, we independently sample 12 responses in thinking mode using
temperature $1.0$, top-$p=0.95$, min-$p=0$, and a maximum generation
length of 38,912 tokens. We report Avg@12. 
\paragraph{Scientific-reasoning evaluation.}
We evaluate the held-out SciKnowEval L3~\citep{feng2024sciknoweval} multiple-choice test splits
for biology, chemistry, physics, and materials science, using the
evaluation hyperparameters of~\citet{hubotter2026reinforcement}.
For each problem, we independently sample 16 responses using
temperature $0.6$, top-$p=0.95$, no top-$k$ filtering, and a maximum
completion length of 8,192 tokens. We report Avg@16. A response is
counted as correct when its extracted final option matches the
ground-truth answer. Evaluation is conducted separately
for biology, chemistry, physics, and materials science, following the
evaluation protocol of~\citet{zhao2026self}, e.g., checkpoint selection, etc.
\subsection{Full Numerical Results for Main-Paper Figures}
This section reports the complete numerical results corresponding to
the ablation figures presented in the main paper. These tables provide
the exact benchmark-level values used to construct Figures~3 and~5.
All experiments use
\texttt{DeepSeek-R1-Distill-Qwen-7B}, and the reported average is
computed across AIME24, AIME25, and HMMT25.

\subsection{Masking-Ratio Results}

Table~\ref{tab:deepseek_masking_ablation} reports the complete numerical
values used to construct Figure~3 of the main paper. The number of
partial reference views is fixed to $K=4$. All evaluated masking ratios outperform the full-reference OPSD
baseline in average accuracy.

\begin{table}[ht]
\centering
\small
\setlength{\tabcolsep}{4pt}
\renewcommand{\arraystretch}{1.08}

\begin{tabular}{lcccc}
\toprule
\textbf{Method}
& \textbf{AIME24}
& \textbf{AIME25}
& \textbf{HMMT25}
& \textbf{Avg.} \\
\midrule
 $\rho=15\%$
& 57.8 & 40.0 & 25.6 & 41.1 \\
$\rho=40\%$
& 55.0 & 43.1 & 25.6 & 41.2 \\
$\rho=50\%$
& 55.6 & 41.9 & 25.3 & 40.9 \\
$\rho=60\%$
& 54.2 & 41.7 & 28.6 & 41.5 \\
$\rho=65\%$
& 55.8 & 41.1 & 26.1 & 41.0 \\
$\rho=70\%$
& 55.8 & 41.9 & 25.0 & 40.9 \\
\bottomrule
\end{tabular}
\caption{Complete numerical results underlying Figure~3 of the main
paper. We vary the masking ratio $\rho$ on
\texttt{DeepSeek-R1-Distill-Qwen-7B} while fixing the number of views
to $K=4$. The average is computed across AIME24, AIME25, and HMMT25.}
\label{tab:deepseek_masking_ablation}
\end{table}
\subsection{Number-of-Views Results}
Table~\ref{tab:deepseek_views_ablation} reports the complete numerical
values used to construct Figure~5 of the main paper. The masking ratio
is fixed to $\rho=40\%$. The complete results confirm the improvement obtained by aggregating
multiple partial views relative to using a single masked view.

\begin{table}[ht]
\centering
\small
\setlength{\tabcolsep}{4pt}
\renewcommand{\arraystretch}{1.08}

\begin{tabular}{lcccc}
\toprule
\textbf{Method}
& \textbf{AIME24}
& \textbf{AIME25}
& \textbf{HMMT25}
& \textbf{Avg.} \\
\midrule
$K=1$
& 54.2 & 40.3 & 26.4& 40.3 \\
$K=4$
& 55.0 & 43.1 & 25.6 & 41.2 \\
$K=7$
& 55.0 & 42.5 & 26.1 & 41.2 \\
$K=9$
& 59.2 & 40.0 & 25.0 & 41.4 \\
\bottomrule
\end{tabular}
\caption{Complete numerical results underlying Figure~5 of the main
paper. We vary the number of partial reference views $K$ on
\texttt{DeepSeek-R1-Distill-Qwen-7B} while fixing the masking ratio
to $\rho=40\%$. The average is computed across AIME24, AIME25, and
HMMT25.}
\label{tab:deepseek_views_ablation}
\end{table}
\vspace{-2 em}
\begin{algorithm}[H]
\caption{Monte Carlo Privileged Distillation}
\label{alg:mcpd}

\begin{algorithmic}[1]

\Require Training dataset $\mathcal{D}$; student policy
$\pi_{\theta}$; frozen teacher $\pi_{\bar{\theta}}$; number of views
$K$; masking ratio $\rho$
\Ensure Updated student parameters $\theta$

\For{each training minibatch $\mathcal{B}\subset\mathcal{D}$}
    \State $\mathcal{L}_{\mathcal{B}} \leftarrow 0$
    \State $N_{\mathcal{B}} \leftarrow 0$

    \For{each $(x,r)\in\mathcal{B}$}
        \State Sample one on-policy trajectory
        $y=(y_1,\ldots,y_T)\sim\pi_{\theta}(\cdot\mid x)$

        \For
            \State Sample an independent masking pattern $\xi_k$
            \State Construct
            $r^{(k)}\leftarrow\operatorname{Mask}(r;\rho,\xi_k)$
            \State Evaluate the frozen teacher along the same trajectory:
            \State \hspace{\algorithmicindent}
            $q_t^{(k)}
            \leftarrow
            \pi_{\bar{\theta}}
            \left(\cdot\mid x,r^{(k)},y_{<t}\right)$
            for $t=1,\ldots,T$
        \EndFor

        \For{each valid, non-padding rollout position $t$}
            \State Form the Monte Carlo marginal target
            $\widehat{q}_{K,t}
            \leftarrow
            \frac{1}{K}\sum_{k=1}^{K}q_t^{(k)}$

            \State Compute the student distribution
            $p_{\theta,t}
            \leftarrow
            \pi_{\theta}(\cdot\mid x,y_{<t})$

            \State Accumulate the distillation loss
            \State \hspace{\algorithmicindent}
            $\mathcal{L}_{\mathcal{B}}
            \leftarrow
            \mathcal{L}_{\mathcal{B}}
            +
            D_{\mathrm{KL}}
            \left(
            \operatorname{sg}[\widehat{q}_{K,t}]
            \,\|\,p_{\theta,t}
            \right)$

            \State $N_{\mathcal{B}}\leftarrow N_{\mathcal{B}}+1$
        \EndFor
    \EndFor

    \State Normalize the loss:
    $\mathcal{L}_{\mathcal{B}}
    \leftarrow
    \mathcal{L}_{\mathcal{B}}/N_{\mathcal{B}}$

    \State Update only $\theta$ using
    $\nabla_{\theta}\mathcal{L}_{\mathcal{B}}$
\EndFor

\Return $\theta$

\end{algorithmic}
\end{algorithm}
\section{MCPD Training Algorithm}

Algorithm~\ref{alg:mcpd} summarizes the MCPD training procedure.
For each example, the student generates one on-policy trajectory from
the problem alone. The frozen teacher evaluates the same student states
under $K$ independently sampled partial views of the reference. The
teacher distributions are averaged in probability space, and only the
student parameters are updated. Here, $\operatorname{sg}[\cdot]$ denotes the stop-gradient operator.
The same student trajectory is shared across all privileged views, so
variation among teacher targets arises only from the partial reference
context.

\end{document}